# When a Tweet is Actually Sexist. A more Comprehensive Classification of Different Online Harassment Categories and The Challenges in NLP

Sima Sharifirad [1] and Stan Matwin[2]
1 Faculty of computer science, Dalhousie University, s.sharifirad@dal.ca
2 Faculty of computer science, Dalhousie University, stan@cs.dal.ca

Abstract

Sexism is very common in social media and makes the boundaries of freedom tighter for feminist and female users. There is still no comprehensive classification of sexism attracting natural language processing techniques. Categorizing sexism in social media in the categories of hostile or benevolent sexism are so general that simply ignores the other types of sexism happening in these media. This paper proposes a more comprehensive and in depth categories of online harassment in social media e.g. twitter into the following categories, "Indirect harassment", "Information threat", "sexual harassment", "Physical harassment" and "Not sexist" and address the challenge of labeling them along with presenting the classification result of the categories. It is preliminary work applying machine learning to learn the concept of sexism and distinguishes itself by looking at more precise categories of sexism in social media.

## 1 Introduction

Language indeed reveals the values of people and their perspectives. There has been a debate about sexist language since the 1960s among female activists [1]. Despite the general concept that sexism is only discrimination about women, there are more concerns about the way women are represented in advertisements, newspapers and social media. There have been many definitions of sexism; for instance, in a more general term in the dictionary, it is defined as a way of discrimination against women [2]. However, sexism is not all about discrimination, and what is happening in social media is far from this definition. There is a need to conceptualize sexism on online platforms differently since its manifestations are different.

Mills [1] provides comprehensive types of definitions of sexism. Some of the definitions pertain to presumed activities associated with women and are secondary to men or traditional and stereotypical beliefs about women e.g. "fuck off and go back to the kitchen, your husband is hungry". Here it is assumed that there are some duties and jobs specifically for a woman and therefore she is classified less as a person in her own right. Sexism seems to be relatively a complex concept which is not easy to define in lexicon level alone and at the same time in the examples. Waseem et al., 2016 [3], collects hateful tweets in the categories of sexist, racist and neither, this work was the first touch of natural language processing and sexism. He considered the eighteen conditions for being hate speech but not bringing specifically definitions for being sexist or racist. In line with the previous research, tweets were categorized into hostile, benevolent and neither by [2]. In their research, hostile sexism tweets are clearly imposing negative emotions and benevolent sexism tweets are indirectly sexist and there are no words of violence in it.

Ruth et al.,2017 [4] considers the tweets towards top feminists and collects a range of online abuse and categorize them into classes such as sexual harassment, physical threats, flaming and trolling, stalking, electronic sabotage, impersonation and defamation. Her research inspired us to come up with new categories. Mills [1] divides sexism into overt sexism and indirect sexism, each with their own subcategories. Inspiring from [4][5], initially, we come up with nine categories. For each category, we come up with the category name, definitions, points, hints and examples. For some categories that were close to each other, we decided to provide some false examples. See below for the categories.

Benevolent sexism (#1)
Definition: Sexist beliefs represent stereotypic beliefs about women and the prescription of restricted roles that are framed in such a way that they can appear to represent positive beliefs or evaluations of women. Points: These tweets are indirectly sexist and do not have any swear words.
Examples: -A wise woman builds her house but a foolish woman tears it down
with her hands.
 - It's less of #adaywithoutwomen and more of a day without feminists, which, to be quite honest, sounds lovely.
 - Lucky is the man who is the first love of a woman, but luckier is the woman who is the last love of a man.

Physical threats (#2)
Definition: attacked in cyberspace on the basis of threats on their female biology. Comments on their physical attributes were made in a violent and aggressive manner that suggested the female body was bad, diseased, disgusting or killing without sexual intention.
Points: classify if it is about the body of female to get a disease, pain, breaks, disgusted, or any violent or aggressive word about body parts or killing without sexual intention.
Examples: -I was told I deserved to die a painful death.
 -how about you get cancer.
 -just putting it out there, you deserve all those deaths you are getting.

Sexual threats (#3)
Definition: harassment of women online as a kind of pornographic invitation? which perpetuates hierarchical gender norms and incites others to display threatening sexual behavior towards the victim. Points: tweets related to forcing women to have sex.
Example: -f..k that cunt, I would with my fist. False example:
 -Hope one of those bitches falls over and breaks her leg (this tweet is more about breaking and falling so its right category is #2)

Body harassment (#4)
Definition: sexual harassment related to the body or beauty of the women in particularly reliant on positioning women as sex objects. Alongside remarks reposted in the hashtag which were concerned with weight, breast size, hotness, lack of hotness or humor of their body or their clothes or beauty.
Example:
 -I cannot stop looking at Nikki's dreadful black crooked bra.

Masculine harassment (#5)
Definition: tweets that are sexism and show the men control over women and relate to maculate or show the inferiority of the women. online abuse which can only be used against the female sex in patriarchal society. Points: tweets against working women, or lack of their physical or mental ability.
Example:
- The menus look like they were made by a 5-year-old little girl...in this case just the mental age of a 5 year old girl I guess - Yes, we get it. You're pretty. Tone down the self promo and just cook

Lack of attractiveness harassment (#6)
Definition: tweets that show the feeling of ignorant or lack of being sexy or
ignorant.
Example: -don't pull the gender card. You're not being harassed because you're a women. It is because you are an ignorant cunt. - stop calling yourselves pretty and hot..you're not and saying it a million times doesn't make you either – why do they keep saying that they are so pretty
Stalking (#7)
Definition: someone sought and compiled information about you and used it to harass, threaten and/or intimidate you. Behaviors include repeated pursuit, premeditation, repetition, and obsession with the victim. Feel continually scared, threatened and vulnerable Points: a women's street address.
Examples: -I have all your names on ledger. 5000$ a month or I torch you all
over the web, forever.
Impersonation (#8)
Definition: treating that the identity will be stolen.
Examples: -if you are seriously gonna try to change gaming, I will hack your account and put gay porn everywhere
General sexist statements (#9)
Definition: the sentences that are not in the previous categories and are insulting and give the sense of violence toward women by using the insulting words.
Points: tweets related to women name calling or words of anger.
Example: -damn girls, you are fine.

After preparing the first instruction, we ran a pilot study of 50 tweets which labelled by one male and 12 women non-activist. Then, we calculate the Kappa score. Considering their feedback and the kappa score made us merge some of the categories into more uniform categories and we came up with the new instruction. We run the study again with the same raters and calculate the Kappa score. The second score was very promising and help us to run the study on the bigger dataset using Crowdflower. All the details of the score and feedbacks are in section 4. To the best of our knowledge, there has not been any previous study considering these categories to identify different types of sexism on online platforms and classifying them using natural language processing.

The goal of the research is to first understand the sexist language and its characteristics in social media attracting natural language processing and to collect the data in more specific categories. The rest of the paper is organized as follows, in the next section we will go through some related works and literature review, in section three we talk about the process of data collection and the hashtags. In section four, we talked about the pilot study, annotation agreement and the feedback. In section five, we talked about the classification algorithms. In section six, we mentioned the challenges of the study and finally the last section, we conclude and talk about the future study.

2 Related work

Introducing sexism as the classification task was first proposed by Waseem, 2016 [3]. He annotated 16 thousand tweets and categorized them as racist, sexist and neither. His dataset is shared in his GitHub with the tweet IDs. However, by the time we downloaded the tweets, most of them were deleted and were no longer available. Out of three thousand tweets collected by Waseem, only two hundred were available at the time of the study. Waseem [3] collected tweets around the famous Australian tv show, My Kitchen Rules, and the hashtag #mkr. We found this hashtag useful but after a closer look we saw that the corpus had a bias and did not contain different types of sexism. He tried different methods such as character level grams and word grams and employed logistic regression with 10-fold cross-validation. In 2017, Akshita and Radhika [2] introduced a new category of benevolent sexism. Categorizing tweets exhibiting sexism if they have one of the three following features: "protective paternalism", "complementary gender differentiation" and "heterosexual intimacy".

They used different types of classifiers and reported the result along with other polarity detection methods. Alessandro et al., [6] tried to automatically detect hate speech, including sexist words along with other types of taboo words by focusing on just words and combinations of nouns, verbs, adjectives and adverbs. Badjatiya et al., [7] took advantage of deep neural networks models and use the Waseem dataset [3]. They used different types of convolutional and recurrent neural networks along with different representation learnings. Long short term memory (LSTM) was combined with random embedding and among them, gradient boosted decision trees had the best performance. Park and Fung, 2017 [8] approached the Waseem dataset and thought of it as a two-step classification: First, if the tweet was abusive or not and second if it was sexist or racist. They proposed a hybrid Convolutional Neural Network (CNN) by combining character level and word level CNN. In another study, Clarke and Grieve, 2017 [9] also used the Waseem dataset. In their study, they proposed a multi-dimensional analysis (MDA) including of being interactive, antagonistic and attitudinal, comparing the classes of sexist and racist tweets. Their study investigates a wide range of linguistic features and texts.

Unlike the previous works focusing on the natural language processing, Ruth et al., 2016 [4] focused on female activists who were very controversial and experienced regular trolling and abuse on Twitter. They reported that about eighty-eight per cent of them were abused on Twitter and only 60 per cent on Facebook. They came up with the top ten categories of abuse as follows: harassment, sexual harassment, threats of physical violence, threats of sexual violence, stalking, flaming and trolling, electronic sabotage, defamation, inciting other to abuse and impersonation. Based on their study, about 40 per cent of their audience experienced sexual harassment and the rest of sexual violence. In another interesting study, Shaw, 2006 [10], collected information of a campaign "Bye Felipe" which feminist posts the screenshot of harassment and sexual entitlement on online dating websites such as Tinder. They collected information from the reactions that women received after implicitly or explicitly saying no to a man.

They argued that women were considered as sex objects. Women mostly experienced rape threats, violent words, facing humorous sexist words and trolling-like participation. In the experiment, there were a lot of good examples of different threats and sexist tweets. Vitis and Gilmour, 2016 [11] brought the definition of Citron [12], mentioning that there are three major components in online harassment including the fact that the victims are women, the harassment is targeted at women and the abuse is in a threatening way. They analyzed different reactions that females had to such harassment. Megarry, 2014 [5] examined the hashtag #mencallmesomething and identified different aspects of sexism on online platforms. They focused on different females from different countries who were harassed on Twitter. They argued that there are different types of online harassment. One of them is related to the concept of masculinity which defines the level of being level headed and women were criticized to be emotionally sensitive. The other form is related to violence against the physical features of women body. This violence can manifest as hatred towards the female body or references toward diseases such as "I hope you get the sexually transmitted disease of vagina cancer". Sexual attacks were another kind of online harassment, such as "fuck that cunt, I would with my first". References to perceived unattractiveness is another prevalent category of online abuse. Death threat such as "tape a plastic bag on her head and kill her live on a webcam" is another category mentioned in [5]. After all these studies, one can come to the conclusion that dividing online harassment into merely hostile and benevolent sexism ignores the more fine-grained categories of sexism on online platforms.

3 Data collection, annotation and preprocessing

We collected data using Twitter search API and used a wide range of tweets and hashtags extracted from the previous studies. There were 290 hashtags including:
"#mkr", "#Everydaysexism", "#instagranniepants", "#mencallmethings", "#mcmt-a", "#gamergate", "#femfreq", "#metoo", "#slutgate", "#Asian drive", "#nigger", "#mkr", "#sjw", "#weareequal", "#womensday" and "#adaywithoutwomen".

The process of collecting data was quite easy using the current hashtags and we also use Latent Dirichlet allocation (LDA) to get the sub hashtags and used them to collect more tweets [13]. List of useful hashtags for collecting sexist tweets shared in author Github1. We filtered tweets that were not in English, removed duplicate tweets, HTML links, words fewer than three characters, and the spam content. However, we didn't remove the hashtags because we found them useful for labelling the tweets, they provided context for the tweets.

**4 Pilot study, annotation agreement and challenges**

We ran our first pilot study using the first instruction comprising the nine categories. We asked one male and 12 females non-activist to label 50 tweets into the nine categories and give us the feedback about the clarity of the instruction, tweets and also the level of task hardship. We calculated the inter-annotator measurement, Fleiss' kappa score [14] because we had more than two raters to calculate the agreement. The score was 0.30, which was a poor score. We then combined the categories into five more general categories based on the feedback and again calculated the Fleiss' kappa score as 0.70 which was a good score for agreement.

There were some challenges and feedback about the initial nine categories. They were some questions about the raters if we want to label the tweets as sexist or if we want to know if the user is sexist. Some tweets needed context to be categorized properly; raters wanted to know if the tweets were the reply to the previous tweet or were the original tweet. Raters wanted to know the gender of the twitter users. Some tweets were not clear if they were toward a male or a female, so they wanted to know who the audience of the tweet was. There were some abbreviations in the tweets that made labelling difficult for them. They found some overlaps between some classes, namely two, four and six in the first instruction.

They found some tweets were generally hateful towards women or feminists and they didn't specifically find them sexist. There was a big difference between what was in their mind about the definition of the sexist tweet and the presented definitions and the categories in the instruction, and it made it interesting for them as well. They found hashtags useful for finding the context of the tweets. They found the hints and false examples useful too. We addressed some of the challenges in our dataset as follows: for clarifying the gender of the twitter users, we used the lexicon provided by Sap et al., 2014 [15] and we get a number of being positive or negative and it identified if the user is female or male. We added some notes and clarified that the task was the multi-class classification of the tweets. When the audience of the tweets were not clear, we asked raters to randomly decide if the tweet was toward women or men. Based on their feedback we

came up with the second instruction:

Indirect harassment (#1):
Definition: These tweets are around 1. stereotypic and traditional believes about women. or 2. show the inferiority of men over women. Or 3. are indirectly sexist and doesn't have any swear words.
Examples: -A wise woman builds her house but a foolish woman tears it down with her hands. -Yes, we get it. You're pretty. Tone down the self promo and just cook. -It's less of #adaywithoutwomen and more of a day without feminists, which to be quite honest, sounds lovely.
Information threat (#2):
Definition: threats toward women that their information will be stolen and revealed or will be misused.
Examples: -if you are seriously gonna try to change gamin, I will hack your account and put gay porn everywhere. - I have all your names and ledger. 5000$ a month or I torch you all over the web, forever.
Sexual harassment (#3):
Definition: these tweets are 1. contain insulting words, name calling, words of anger and violence.2. force toward sex.
Examples: -damn girls, you are fine. -f..k that cunt, I would with my fist.
Physical Harassment (#4):
Definition: these tweets are about 1. Threats on female biology attributes, beauty of women.2. considering women as se objects.3. implying sense of ignorance or lack of attractiveness. 4. Implying lack of physical or mental ability of women.
5. Using humour toward the female body. Tweets can contain insulting words or violence, but the concentration is on the previous 5 points.
Hint: female body getting disease, pain, breaks, disgusted. Comments on weight,
breast size, hotness, lack of hotness.
Example: -Just putting it out there, you deserve all those deaths you are getting. -don't pull the gender card. You are not being harassed because you are a woman. It is because you are an ignorant cunt. -the menus look like they were made by a 5 year-old little girl. In this case just the mental age of a 5-year-old girl I guess.
Not sexist (#5):
These tweets are not in the previous categories and are not sexist.

## 5 Technical aspect

After coming up with the right instruction, we collected approximately three thousand tweets and 150 test questions, or in other words gold questions. The 150 test questions were useful for testing the performance of the raters and increasing the accuracy of their work. These 150 questions were labelled by the author and another female non-activist. We used Crowdflower [16], which assigns each rater a trusted score based on the number of correct answers on gold questions. For instance, for the one hundred tweets and eight gold questions, If the raters answered eight out of eight questions, they can label batches of 20, if they answer 7 out of eight questions, batches of 15, 6 out of eight questions, batches of 10. Table1 shows the amount of data in each category.

| Table 1. Classification results on accuracy | | |
|---|---|---|
| Name of the Categories | Number in each class | Data Rater's confidence score |
| Indirect harassment(#1) | 260 | 0.75 |
| Information threat(#2) | 2 | 0.77 |
| Sexual harassment(#3) | 417 | 0.75 |
| Not sexist(#5) | 2440 | 0.75 |

After getting the labelled data, we ran our experiment using a wide range of representation learning, neural networks and Naive Bayes as the classifier. The result of these calculations reported in Table 2. We used Long short term memory (LSTM) using Keras, 8 batches of size 30 each, sequence-length of 15, the learning rate of 0.001 and size of 128. For character level Convolutional Neural network, we used Tensorflow and the code presented in [17].

| Table 2. Classification results on accuracy. | |
|---|---|
| Methods | Algorithm(NB) |
| bigrams(BOW) | 0.73 |
| threegrams(BOW) | 0.68 |
| four grams(BOW) | 0.66 |
| Two character grams(BOW) | 0.79 |
| Three character grams(BOW) | 0.80 |
| Four character gram(BOW) | 0.79 |
| combination of all grams | 0.73 |
| combination of all character-grams | 0.81 |
| combination of all character-grams and all the grams | 0.75 |
| Word2vec | 0.79 |
| Doc2vec | 0.77 |
| LSTM | 0.91 |

## 6 Challenges of study

Collecting the data, labeling and working on the sexism as the working dataset made people look slightly different on you. Their first idea is that they author is a feminist or a girl who witnessed a lot of inequality in her life. Some have reactions of humour, ridicule and irony toward this topic of research. It is also challenging if one can see the sexist sentences as hateful type of speech. They both share some characteristics such as violence and threats. The other challenge is defining the boundaries between the hate speech and sexism tweets. Looking into the tweets, some are just hateful toward feminist.

## 7 Conclusion

Sexism is a very prevalent in social media. In this study, we present new categories of sexism on social media and address the challenges of coming up with the right categories, right annotation and right instruction. Furthermore, we report our initial classification report on these categories. Our future studies will be considering to collect more data base on the categories, investigating more classification tasks and deploying better representation learnings.